\documentclass{article}

\PassOptionsToPackage{numbers, compress}{natbib}

\usepackage[preprint]{neurips_2026}

\usepackage[utf8]{inputenc} 
\usepackage[T1]{fontenc}    
\usepackage{hyperref}       
\usepackage{url}            
\usepackage{booktabs}       
\usepackage{amsfonts}       
\usepackage{nicefrac}       
\usepackage{microtype}      
\usepackage{xcolor}         

\definecolor{darkblue}{rgb}{0, 0, 0.5}
\hypersetup{
    colorlinks=true,
    citecolor=darkblue,
    linkcolor=darkblue,
    urlcolor=darkblue
}
\hypersetup{
    colorlinks=true,
    citecolor=blue,
    linkcolor=blue,
    urlcolor=blue
}

\usepackage{enumitem}       
\usepackage{subfig}         
\usepackage{multirow}       %
\usepackage{xspace}         
\usepackage{wrapfig}        
\usepackage{makecell}       
\usepackage{graphicx}       
\usepackage{colortbl}
\usepackage{bbm}            
\usepackage{adjustbox}

\usepackage{bm}
\usepackage{pifont}         
\usepackage{bbding}         
\usepackage{xurl}           
\NewDocumentCommand{\drsh}{ O{0.6em} O{0.5em} O{0.65pt} O{rounded corners=1.6pt} }{%
  \mathrel{%
  \hspace{0.3em}
    \tikz[baseline=-0.6em]{%
      \draw[->, line width=#3, #4] (0,0) -- (0,-#2) -- (#1,-#2);
    }%
  }%
}

\usepackage{amsmath}
\usepackage{amssymb}
\usepackage{mathtools}
\usepackage{amsthm}

\usepackage[capitalize,noabbrev]{cleveref}

\theoremstyle{plain}
\newtheorem{theorem}{Theorem}[section]
\newtheorem{proposition}[theorem]{Proposition}

\theoremstyle{definition}

\theoremstyle{remark}

\usepackage[textsize=tiny]{todonotes}

\usepackage{algorithm}
\usepackage{algorithmicx}
\usepackage{algpseudocode}

\usepackage{array}
\newcolumntype{N}{c@{\hspace{2pt}}}    
\newcolumntype{L}[1]{>{\raggedright\arraybackslash}p{#1}}
\newcolumntype{C}[1]{>{\centering\arraybackslash}p{#1}}

\setlist[itemize]{leftmargin=20pt}
\setlist[enumerate]{leftmargin=20pt}

\usepackage[most,skins,theorems,breakable]{tcolorbox}
\tcbset{
  mybox/.style={
    width=\linewidth,
    top=14pt,
    bottom=7pt,
    colback=cyan!5!white,
    colframe=black,
    colbacktitle=black,
    enhanced,
    center,
    breakable,
    attach boxed title to top left={yshift=-0.13in,xshift=0.15in},
    boxed title style={boxrule=0pt,colframe=white,},
  }
}
\newtcolorbox{mybox}[2][]{mybox,title=#2,#1}

\newcommand{\ourmethod}{ASPO\xspace}

\newcommand{\mycolor}{cyan!10}

\newcommand{\DSOneB}{DeepSeek-R1-Distill-Qwen-1.5B\xspace}
\newcommand{\QFourB}{Qwen3-4B\xspace}
\newcommand{\QEightB}{Qwen3-8B\xspace}
\newcommand{\QThirtyB}{Qwen3-30B-A3B\xspace}

\usepackage{circledsteps}

\newcommand\blfootnote[1]{%
  \begingroup
  \renewcommand\thefootnote{}\footnote{#1}%
  \addtocounter{footnote}{-1}%
  \endgroup
}

\let\citet\cite

\title{%
When Importance Sampling Misallocates Credit: Asymmetric Ratios for Outcome-Supervised RL%
}

\author{%
  Jiakang Wang$^{2*}$, Runze Liu$^{1,2,3*}$, Qingpeng Cai$^{2}$, Lei Lin$^{2}$, Wenping Hu$^{2}$, \\
  {\bf Xiu Li$^{3}$, Fuzheng Zhang$^{2}$, Guorui Zhou$^{2}$, Kun Gai$^{2}$, Ling Pan$^{1}$} \\
  $^{1}$The Hong Kong University of Science and Technology, $^{2}$Kuaishou Technology, \\ $^{3}$Tsinghua University
}

\begin{document}

\blfootnote{$^*$ Equal contribution}

\maketitle

\begin{abstract}
Reinforcement learning (RL) has shown great promise in large language models (LLMs) post-training, which typically rely on token-level clipping to maintain stability during optimization. Despite the empirical success of GRPO-style methods, we identify a fundamental and previously overlooked challenge in this popular Outcome-Supervised RL (OSRL) paradigm. We reveal that in OSRL, where advantages are shared across tokens within a response, \textbf{importance sampling (IS) ratios deviate from their traditional purpose of distribution correction as in classic RL, which become token-level weights} that allocate the shared advantage signal across tokens. We show that this hidden role shift induces a critical mismatch for positive-advantage tokens, leading to unbalanced token weighting between positive and negative tokens. Specifically, it suppresses the update of underrepresented tokens that are lagging behind, while over-amplifying already high-probability tokens. This mismatch results in rich-get-richer dynamics that over-reinforce confident tokens, weaken catch-up learning that drive entropy collapse, excessive repetition, and premature convergence. To address this, we propose \textbf{A}symmetric Importance \textbf{S}ampling \textbf{P}olicy \textbf{O}ptimization (ASPO), a simple yet effective strategy that reverses the ratio-induced weighting of positive-advantage tokens, while stabilizing extreme updates and maintaining gradient flow. This mismatch correction aligns their update direction with the learning dynamics of negative ones. Comprehensive experiments across math reasoning and coding benchmarks demonstrate that ASPO significantly mitigates entropy collapse, improves training stability, and enhances performance over strong GRPO-based baselines. Our analysis provides new insights into the role of token-level weighting in OSRL and highlights the critical importance of correcting ratio-induced weighting in LLM RL.
\end{abstract}

\section{Introduction}
\label{sec:introduction}

Reinforcement Learning~(RL) has achieved remarkable success across various domains, including board games~\citep{AlphaGo-Zero, AlphaZero}, robotic control~\citep{kober2013reinforcement, andrychowicz2020learning, Bi-DexHands}, Large Language Model~(LLM) alignment~\citep{ouyang2022training, bai2022training} and reasoning~\citep{DeepSeek-R1, o1, DAPO}. In classical RL, an agent interacts with an environment and receives intermediate rewards at different timesteps along a trajectory, where different actions contribute unequally to the final return, leading to fine-grained credit assignment for effective policy learning.

\begin{figure}[!h]
\centering
\includegraphics[width=0.8\textwidth]{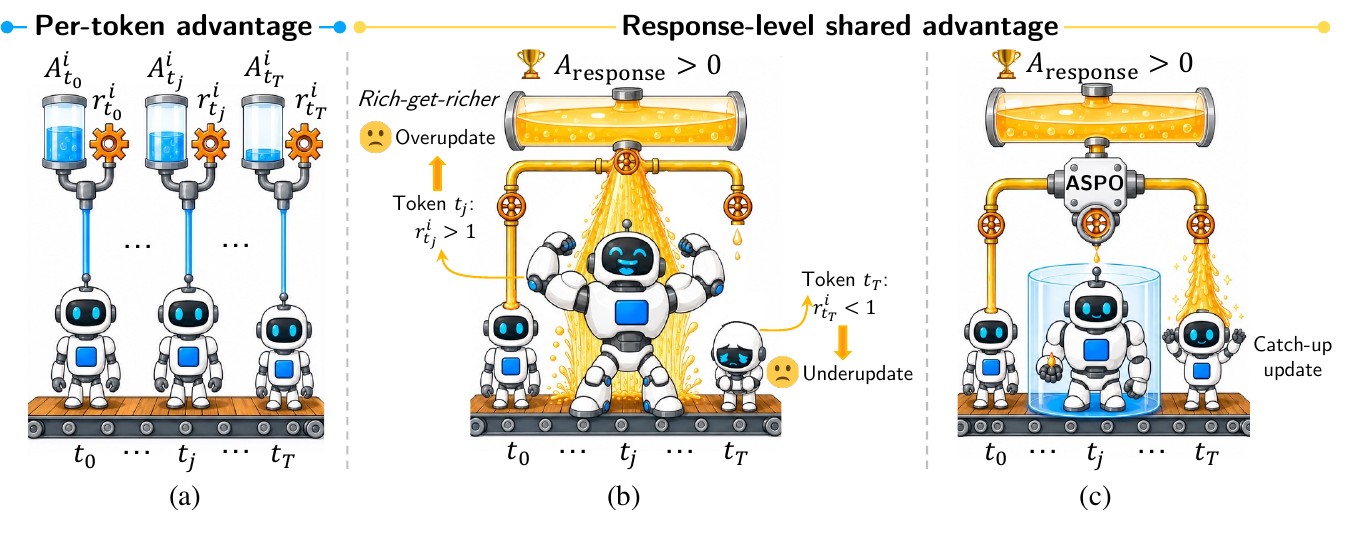}
\caption{Illustration of ratio-induced token misallocation in outcome-supervised RL and the ASPO correction.
For positive tokens, \ourmethod prevents overoptimization of high-probability tokens and facilitates low-probability token learning.}
\label{fig:diagram}
\end{figure}

In this regime, policy optimization methods such as Proximal Policy Optimization (PPO)~\citep{PPO} rely on importance sampling (IS) ratios to account for the distribution mismatch between the behavior policy that generated the data and the current policy being optimized. Together with the clipping mechanism, IS plays a principled role in reusing on-policy samples for optimization while preventing excessively large policy changes.
In standard PPO, the IS ratio is coupled with a per-step advantage $A_t$, and each action-specific ratio modulates the contribution of the corresponding action-specific advantage (Figure~\ref{fig:diagram}(a)).
Thus, the probability ratio primarily serves as a distribution-correction factor within a fine-grained credit-assignment structure.

Outcome-Supervised Reinforcement Learning~(OSRL) for LLMs departs sharply from the traditional RL paradigm~\citep{zhang2025survey}. Group Relative Policy Optimization (GRPO)~\citep{GRPO} and its variants, such as DAPO~\citep{DAPO}, have become widely adopted in this setting due to their simplicity and effectiveness. They estimate a response-level advantage from group rollouts $\hat{A}_{\mathrm{resp}}$ which is \emph{shared} across all tokens in the response. Despite its empirical success, GRPO inherits optimization machinery from PPO (e.g., IS ratios) in a regime that is substantially different from the one for which PPO was originally designed.
We uncover a previously underexplored consequence of this design: under the commonly-adopted shared advantage setups with $\hat{A}_{\mathrm{resp}}$ in OSRL, the standard distribution-correction of PPO-style ratios becomes incomplete.
Since $\hat{A}_{\rm resp}$ is shared across all tokens, it no longer provides token-level specific credit assignment, and consequently, the ratio degenerates into intra-response \textbf{\emph{token-level weights}} that determine how strongly each token contributes to the gradient update.

This role shift leads to a critical mismatch in token weighting, as shown in Figure~\ref{fig:is_visualization}. When the old-policy probability is fixed, for positive-advantage samples, tokens that already have higher probability under the current policy receive larger update weights, while tokens with low probability are heavily down-weighted.
This creates a richer-get-rich dynamic: already confident tokens in successful responses are reinforced more aggressively, while lagging but desirable tokens are under-updated when they may need stronger learning signals to catch up, as illustrated in Figure~\ref{fig:diagram}(b), which has been overlooked in previous research.
Existing modifications to GRPO, such as clipping-threshold adjustments in DAPO, partially relax the upper clipping bound to facilitate low-probability token learning, but does not address the core issue that low-probability positive tokens remain weakly updated. 
This design contradicts the desired learning dynamics, where low-probability tokens in positive trajectories should receive stronger updates to catch up, and result in a series of abnormal training dynamics.
Empirically, we observe accelerated entropy collapse, increasing repetition rate, rising clipping ratios, and premature convergence (Figure~\ref{fig:main_results}).
These phenomena indicate unstable optimization driven by self-reinforcing updates on confident tokens, rather than healthy convergence.

To address these issues, we propose \textbf{A}symmetric Importance \textbf{S}ampling \textbf{P}olicy \textbf{O}ptimization (\ourmethod), a simple yet effective method for GRPO-style OSRL.
\ourmethod is based on the observation that, once a shared response-level advantage is broadcast to all tokens, the ratio primarily determines how this outcome signal is distributed across tokens. For positive-advantage responses, ASPO reverses the direction of ratio-induced token weighting: low-probability positive tokens receive stronger updates, while already confident tokens are down-weighted. This directly counteracts the rich-get-richer effect of standard GRPO, as shown in Figure~\ref{fig:diagram}.
To further stabilize training, we incorporate a soft dual-clipping mechanism~\citep{Dual-clip-PPO, CISPO} that constrains extreme ratios without discarding gradients for positive tokens.
Extensive experiments on both mathematical reasoning and coding benchmarks demonstrate that \ourmethod:
(1) mitigates entropy collapse and overfitting,
(2) yields more stable training dynamics, and
(3) significantly outperforms GRPO-based baselines in final performance.

In summary, our main contributions are as follows:
\begin{itemize}[leftmargin=*, itemsep=0pt, topsep=0pt, parsep=0pt, partopsep=0pt]
    \item We identify a role shift of IS ratios in widely-adopted OSRL and a fundamental misallocation, where the IS ratio acts as token-level weighting and can misallocate credit across tokens: \emph{positive-advantage tokens are weighted in the undesired direction}, with confident tokens overoptimized and lagging tokens suppressed.
    \item We propose \ourmethod, an asymmetric ratio-based mechanism that promotes catch-up learning for underrepresented successful tokens while preserving gradient flow.
    \item We provide extensive empirical evidence that correcting token-level weighting significantly improves training stability and performance across multiple math reasoning and coding benchmarks.
\end{itemize}

\section{Related Work}
\label{sec:related_work}

\noindent \textbf{RL for Large Language Models.}
Reinforcement Learning from Human Feedback (RLHF)~\citep{christiano2017deep, MRN} has achieved remarkable success in aligning LLMs with human values. Recently, DeepSeek-R1~\citep{GRPO} and the GRPO algorithm~\citep{GRPO} have demonstrated that RLVR effectively enhances the reasoning capabilities of LLMs. Subsequent works have extended the GRPO algorithm for RLVR~\citep{DeepScaleR, DAPO, VAPO, Skywork-OR1, Archer, AttnRL, GPPO, CE-GPPO, FlowRL}. Our method follows this GRPO-based line of research but introduces a new approach of the clipping and IS mechanism to address inherent limitations in GRPO.

\noindent \textbf{Clipping Mechanism in RL.}
PPO~\citep{PPO} introduces clipping based on importance sampling ratios as an alternative to KL divergence to constrain policy updates relative to the reference policy, and GRPO~\citep{GRPO} adopts this clipping loss for LLM RL. There has been a growing line of work investigating whether this clipping is the right stabilization mechanism for LLM RL. CISPO~\citep{CISPO} observes that gradients for clipped tokens are masked and proposes preserving them based on the PPO-Clip objective~\citep{PPO} but without the conservative $\min$ operation. GSPO~\citep{GSPO} argues that the optimization objective should match the sequence-level reward's granularity, proposing sequence-level clipping and IS ratio computation. DCPO~\citep{DCPO} employs dynamic-adaptive clipping ranges instead of fixed bounds. Our method also targets the clipping term in GRPO but differs primarily by focusing on the IS ratio, incorporating a reciprocal weight for positive tokens. DAPO~\citep{DAPO} and DPPO~\citep{DPPO} further relax the clipping ranges of low-probability (positive) tokens, but do not address the problem of weight mismatch of positive tokens. These studies mainly refine the gating behavior of ratio-based objectives, deciding when updates should be clipped, masked, relaxed, or computed at a different granularity to maintain training stability. 
In contrast, our work focuses on the overlooked phenomenon of relative weighting of updates that remain active.

\section{Preliminaries}
\label{sec:preliminaries}

\subsection{Group Relative Policy Optimization}
Group Relative Policy Optimization~(GRPO)~\citep{GRPO} samples a group of $G$ rollouts for advantage estimation: $\hat{A}_t^i = \frac{R^i - \operatorname{mean}(\{R^i\}_{i=1}^G)}{\operatorname{std}(\{R^i\}_{i=1}^G)}$. The loss function of GRPO is defined as:
\begin{equation}
\begin{aligned}
\mathcal{J}_{\text{GRPO}}(\theta) = &\ \mathbb{E}_{q \sim \mathcal{D}, \{o^i\}_{i=1}^G \sim \pi_{\theta_{\text{old}}}(\cdot \mid q)} \\
& \left[ \frac{1}{G} \sum_{i=1}^G \frac{1}{|o^i|} \sum_{t=1}^{|o^i|} \bigg( \min\left( r_t^i(\theta) \hat{A}_t^i, \operatorname{clip}\big( r_t^i(\theta), 1-\varepsilon, 1+\varepsilon \big) \hat{A}_t^i \right) - \beta \mathbb{D}_{\text{KL}}(\pi_\theta \| \pi_{\text{ref}}) \bigg) \right],
\end{aligned}
\label{eq:grpo_loss}
\end{equation}
where $r_t^i = \frac{\pi_{\theta}(o_t^i \mid q, o_{<t}^i)}{\pi_{\theta_{\text{old}}}(o_t^i \mid q, o_{<t}^i)}$ is the Importance Sampling (IS) ratio, and $\beta$ is a weight for the Kullback-Leibler~(KL) divergence between the current policy $\pi_\theta$ and the reference policy $\pi_{\text{ref}}$.

\subsection{PPO Clipping and the Improvements}

The clipping mechanism in GRPO~\citep{GRPO} is first introduced by PPO-Clip~\citep{PPO}. It serves as an simple yet effective constraint to prevent large policy updates. However, it is shown in \citet{CISPO} that this clipping mechanism clips the value and also masks the gradient of the clipped tokens.

\section{The Role Shift of Importance Sampling Ratio in OSRL}
\label{sec:is_not_important}

\subsection{Theoretical Understanding: PPO-style IS Ratios as Induced Token Weights in OSRL}
\label{subsec:is_not_important_motivation}

Importance Sampling (IS) enters policy optimization as a way to evaluate an objective under one policy using samples collected from another~\citep{Importance-Sampling} for optimizing $\sum_a \pi_\theta(a|s) A_{\theta_{\mathrm{old}}}(s,a)$, where $\theta$ and $\theta_{\rm old}$ denote parameters of the current and old policies, respectively, with $A$ the stepwise advantage function for the current step's state action pair $(s,a)$.
Since trajectories are sampled from the old policy $\pi_{\theta_{\mathrm{old}}}$, this expectation is rewritten using an importance sampling estimator $\mathbb{E}_{a\sim \pi_{\theta_{\mathrm{old}}}(\cdot|s)}
\left[
\frac{\pi_\theta(a|s)}
{\pi_{\theta_{\mathrm{old}}}(a|s)}
A_{\theta_{\mathrm{old}}}(s,a)
\right]$.
This derivation gives the likelihood ratio $r_t=\frac{\pi_{\theta}}{\pi_{\rm old}}$ a clear interpretation, which corrects the distribution mismatch between the old policy that generated the data and the current policy being optimized for that action~\citep{PPO}. In the classic PPO surrogate in the RL literature, this IS ratio is coupled with a step-specific advantage $A_t$, which specifies the estimated credit of the same timestep $t$ (operating at the same granularity).

In OSRL for LLMs, such as GRPO~\citep{GRPO} and its variants (e.g., DAPO~\citep{DAPO}), this granularity alignment is broken. GRPO-style methods still use token-level probability ratios $r_t$, but the advantage is not estimated at the token level. Instead, a response-level advantage $\hat{A}_{\rm resp}$ is estimated and then broadcast to every token in the response, i.e., $A_t^i = \hat{A}^i, \forall t$. Thus, all tokens within a response \emph{share the same advantage value}, which is fundamentally different from the classical setting where $A_t$ provides timestep-specific credit. 
Once the same $\hat{A}_{\rm resp}$ is assigned to every token, the advantage can no longer distinguish which tokens should receive stronger or weaker updates. 

The original motivation of IS is to correct the distribution mismatch of the action-specific advantage term. However, under shared advantages, there is no token-specific credit estimation, which makes it unclear what the token-level ratio is correcting when the advantage itself no longer provides token-level credit, leading to the following critical question:
\begin{center}
\textit{
If the real credit of each token is already unclear due to outcome-based advantage estimation, what is the role and effect of further adjusting the distribution using IS weights?
}
\end{center}

We first answer this question theoretically by analyzing the gradient of GRPO, which is further empirically validated in Section~\ref{subsec:is_not_important_exp}. The following proposition shows that it is gradient-equivalent to a weighted log-likelihood objective with ratio-induced token weights.

\begin{proposition}[IS ratio as token-wise gradient weighting in GRPO]
Consider the GRPO objective in Eq.~(\ref{eq:grpo_loss}). For each token $o_t^i$, define the active clipping mask $m_t^i=\begin{cases}1, & A_t^i \ge 0 {\rm{\ and\ }} r_t^i \le 1+\epsilon, \\
1, & A_t^i < 0 {\rm{\ and\ }} r_t^i \ge 1-\epsilon, \\
0, & {\rm{otherwise}}
\end{cases}$
and the detached token weight $w_t^i={\rm{sg}}(m_t^i r_t^i)$, where $\rm{sg}(\cdot)$ denotes the stop-gradient operator. Consider the weighted log-likelihood objective in Eq.~(\ref{eq:wll_loss}), then we have $\nabla_\theta \mathcal J_{\mathrm{GRPO}}(\theta)=\nabla_\theta\mathcal J_{\mathrm{WLL}}(\theta)$.
\begin{equation}
\mathcal{J}_{\text{WLL}}(\theta) = \mathbb{E}_{q \sim \mathcal{D}, \{o^i\}_{i=1}^G \sim \pi_{\theta_{\text{old}}}(\cdot \mid q)} 
\left[ \frac{1}{G} \sum_{i=1}^G \frac{1}{|o^i|} \sum_{t=1}^{|o^i|} \bigg( w_t^i A_t^i \log \pi_{\theta}(o_t^i | q,o_{<t}^i) - \beta \mathbb{D}_{\text{KL}}(\pi_\theta \| \pi_{\text{ref}}) \bigg) \right].
\label{eq:wll_loss}
\end{equation}
\label{prop}
\end{proposition}
\noindent \textbf{Remark.} Proposition~\ref{prop} reveals how the IS ratio enters the GRPO update, and the proof can be found in Appendix~\ref{sec:proof}. 
Because GRPO computes the advantage at the response level and broadcasts it to all tokens, the token-level ratio $r_t^i$ is no longer paired with a token-specific credit estimate.
Instead, after clipping, it becomes the main token-dependent scalar that modulates the update of each token on the shared advantage, i.e., ${\rm sg}(m_t^i r_t^i) \hat A^i$.
Therefore, tokens within the same response can receive different update magnitudes solely because their current-to-old probability ratios differ, rather than because their individual contributions to the response reward are different.

\subsection{Experimental Validation}
\label{subsec:is_not_important_exp}

The gradient view in Appendix~\ref{sec:gradient_analysis} motivates our empirical analysis.
Although IS ratios are introduced to compensate for the mismatch between the behavior policy and the current policy, GRPO-style OSRL computes advantages at the response level and broadcasts them to all tokens. 
Thus, the IS ratio becomes the main token-specific factor in the effective update weight. 
We therefore ask whether IS is necessary as a distribution-correction mechanism in this setting, or whether it mainly reshapes learning dynamics through ratio-induced weighting.

\paragraph{Setup.}
To evaluate the practical effects of IS weights in OSRL, we compare two variants that are identical except for the use of $r_t$ ratios:
(1) GRPO with original IS weights, and (2) GRPO without IS weights (where all IS weights are fixed to 1.0).
If IS were indispensable for correcting this mismatch, removing it would be expected to noticeably degrade performance or destabilize training.

\subsubsection{Results}
\begin{figure*}[!t]
\centering
\includegraphics[width=1\linewidth]{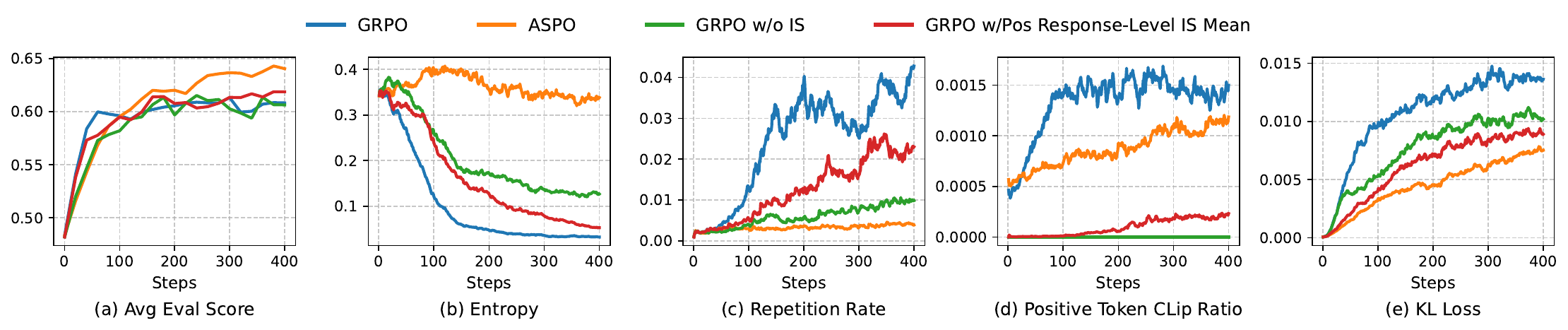}
\caption{The training dynamics curves of all methods on \QFourB, including (a) average evaluation accuracy, (b) entropy, (c) repetition rate, (d) positive token clip ratio, and (e) KL loss. The curves are smoothed with EMA for better visualization.}
\label{fig:main_results}
\end{figure*}

Figure~\ref{fig:main_results} compares the evaluation accuracy and training dynamics of standard GRPO and GRPO without IS.
The two variants achieve comparable final test accuracy.
Standard GRPO reaches its peak slightly earlier and stops to increase, whereas GRPO without IS converges more smoothly with competitive performance.
Its training dynamics are also more stable, with slower entropy decay and smaller increases in repetition, and KL divergence.
These results suggest that removing IS does not compromise final performance, but makes optimization less aggressive.

\subsubsection{Analysis}
\begin{figure*}[!h]
\centering
\includegraphics[width=0.8\textwidth]{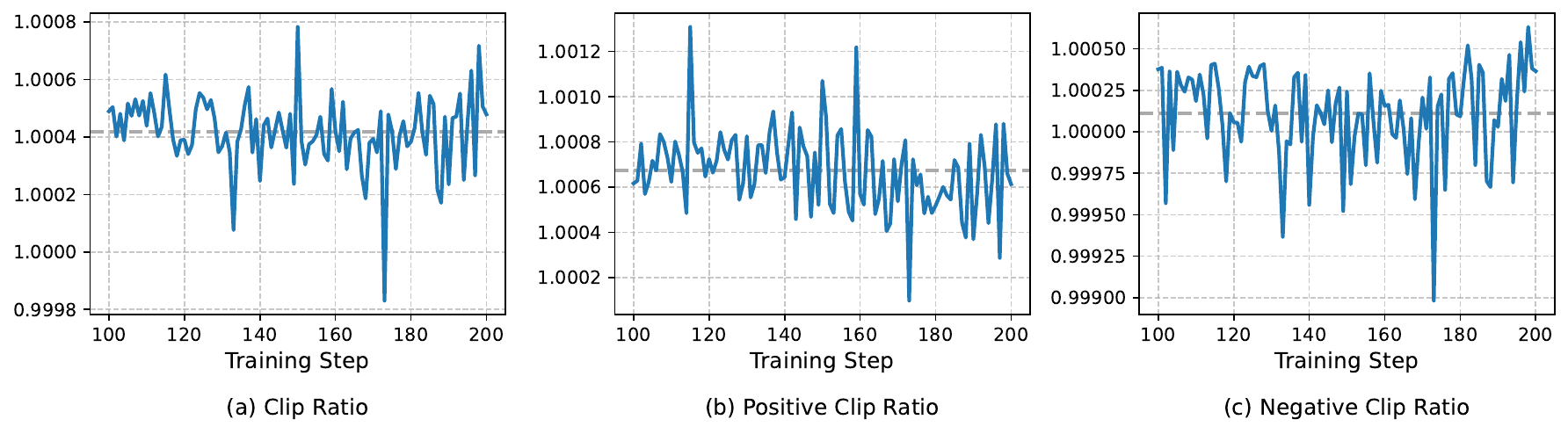}
\caption{Curves of response-level IS ratios throughout GRPO training. The average IS ratios are shown in gray dashed lines.}
\label{fig:response_level_is_ratio}
\end{figure*}

To understand what the IS ratios are doing, we further examine their values during GRPO training.
Figure~\ref{fig:response_level_is_ratio} shows the response-level IS ratios, where the average IS ratio is typically slightly greater than 1.0 (around 1.0004), and the ratios are not uniformly distributed across samples. Specifically, the average IS ratio of positive samples is slightly higher than that of negative samples. While the numerical gap (for each update) is small, it accumulates over many tokens and repeated off-policy updates, and its cumulative effect is significant.
According to Proposition~\ref{prop}, the GRPO gradient can be written as a weighted log-likelihood update whose effective weight is proportional to $r_{i,t}\hat{A}_i$, which changes the relative strength of the corresponding gradient terms. For positive responses, ratios above one amplify the reward-increasing update. For negative responses, ratios below one tend to shrink the magnitude of the penalizing update in the gradient-active region.
Thus, it causes GRPO training to \textbf{prioritize learning positive responses rather than suppressing negative ones}, and this gap widens across training, which amplifies positive-sample learning and accelerates entropy decay.

\paragraph{What happens without IS weights?}
With all IS ratios set to 1.0, the update no longer contains this ratio-induced preference amplification. 
On the one hand, this makes the overall weights slightly lower than in standard GRPO. On the other hand, it \emph{eliminates the weight difference between positive and negative samples}. While these two factors slow down the learning speed compared to standard GRPO, they do not compromise the final performance, as shown in Figure~\ref{fig:main_results}.

This observation supports the role-shift interpretation in Section~\ref{subsec:is_not_important_motivation}.
In GRPO-style OSRL, although IS ratios still quantify the deviation between the current policy and the behavior policy, when a response-level advantage is shared by all tokens, their dominant operational effect is to reallocate the outcome-level learning signal across samples and tokens.
The more aggressive entropy decay, repetition growth, clipping, and KL increase observed in Figure~\ref{fig:main_results} are consistent with this ratio-induced amplification effect, and motivate the finer-grained token-level analysis in the next section.

Based on the above analysis, we draw the following conclusions:
\begin{mybox}{Takeaways for Importance Sampling}
\begin{itemize}[leftmargin=6pt]
    \item In the shared-advantage GRPO setting, removing ratio-induced weights preserves comparable final performance, while making optimization smoother and less aggressive. 
    \item Although IS ratios still measure current-to-old policy deviation, their dominant operational role is to act as effective update weights under shared response-level advantages.
    \item IS ratios induce an asymmetric weighting pattern: positive responses are amplified more than negative responses are suppressed.
\end{itemize}
\end{mybox}

\section{Ratio-induced Weight Mismatch in Positive-Advantage Responses}
\label{sec:is_mismatch}

\subsection{Rethinking the Role of IS-induced Token Weights}
The analysis in Section~\ref{sec:is_not_important} shows that, under shared response-level advantages, PPO-style IS ratios (after clipping) in GRPO-style OSRL primarily act as \textbf{token-level training weights}, which modulate how strongly each token is updated under the same response-level advantage. 
This motivates a different question: \emph{if the ratio is viewed as an update weight, does the PPO-Clip weighting pattern allocate learning signal to tokens in a desirable way?}

\begin{figure*}[!h]
\centering
\includegraphics[width=1\textwidth]{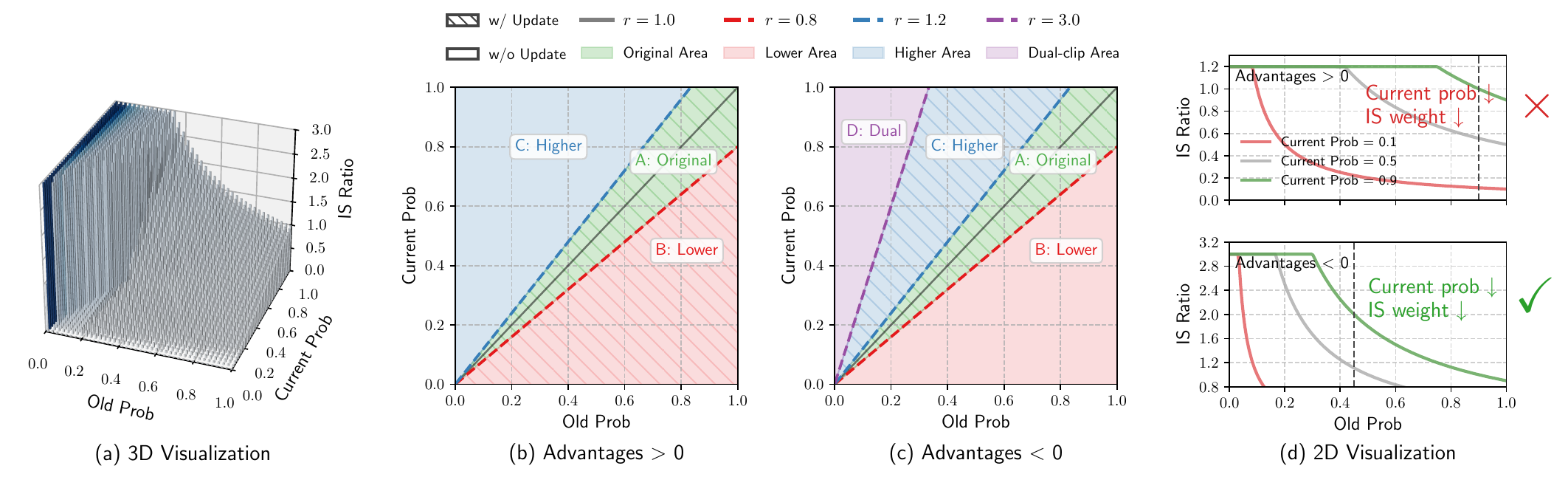}
\caption{Visualization of IS weights in PPO-Clip.
(a) 3D visualization of IS weights, showing that for a fixed old probability, tokens with lower current probabilities receive smaller IS weights.
(b) When the advantage is positive ($A > 0$), GRPO optimizes regions A and B.
(c) When the advantage is negative ($A < 0$), GRPO optimizes regions A and C.
(d) 2D visualization of IS weights, illustrating that along a fixed old probability (dashed line), IS weights monotonically decrease as the current probability decreases. For negative-advantage tokens, this behavior aligns with our expectations, whereas for positive-advantage tokens, it contradicts the desired optimization trend.}
\label{fig:is_visualization}
\end{figure*}

The design principle of PPO-Clip is to ensure training stability by preventing tokens that already have a strong advantage in the update direction from dominating the update. This avoids overly aggressive parameter changes that could push the model too far from the old policy.
An ideal weighting scheme might look like this: along the update direction of the advantage, the lower a token's probability is relative to the old policy, the larger its training weight should be. Conversely, the higher its probability, the smaller the weight should be. There are two reasons for this:
(1) Assigning higher weight to tokens that are lagging behind accelerates their learning progress.
(2) Such tokens are already far from the old policy, so updates pose less risk of destabilization.
To make this more intuitive, we visualize IS weights in a three-dimensional coordinate plot, where the z-axis represents the original IS ratio, as shown in Figure~\ref{fig:is_visualization}(a). It can be seen that when the current probability is low, the corresponding IS weight is also small, resulting in insufficient training for these tokens.

\subsection{Weight Misallocation of Positive Tokens in PPO-Clip}
\label{subsec:is_mismatch_positive}

As shown in Figure~\ref{fig:is_visualization}(c), for negative-advantage samples, the weight distribution behaves as expected: weights decrease gradually from the top-left region to the bottom-right region. However, for positive-advantage samples shown in Figure~\ref{fig:is_visualization}(b), the allocation is the opposite to our intuition. Tokens in the top-left region, whose probabilities under the current policy are already much larger than under the old policy, are given larger weights, while tokens in the bottom-right region, with lower current probabilities, are assigned very small weights. Therefore, PPO-Clip underweights lagging tokens and overweights already reinforced tokens within positive-advantage samples.

This mismatch causes two problems: (1) tokens in the bottom-right region, which are clearly lagging behind the old policy, are suppressed further by excessively low weights. For example, if the old policy probability is $0.9$ and the current policy probability is $0.1$, the assigned weight is merely $1/9$, resulting in a weak update signal and insufficient learning.
(2) for tokens in the top-left region that already have a significant advantage, they are assigned disproportionately high weights. On the one hand, the model is more likely to deviate from the old policy, undermining training stability.
On the other hand, the model overfits on positive samples, further amplifying their probabilities after the update. In the next update, their weights increase even more, forming a \textbf{self-reinforcing loop}. This mechanism underlies the previously observed phenomena such as entropy collapse and increased output repetition.
For more details on how to distinguish healthy convergence and local optima, please refer to Appendix~\ref{app:distinguish_convergence_or_local_optima}.

\subsection{Experimental Validation}

\paragraph{Setup.}
To empirically validate the preceding analysis, we conduct a controlled comparison between the original GRPO baseline and a modified variant. In this variant, we replace the token-level IS ratios of \textbf{positive-advantage samples} with their \textbf{response-level average IS ratios}, while keeping the negative-advantage samples unchanged. This design isolates the impact of the mismatched IS weights identified in Section~\ref{subsec:is_mismatch_positive}, ensuring that any observed behavioral differences arise solely from the reweighting of positive tokens.
If our hypothesis is correct, the modified setup should partially alleviate the weight mismatch issue.

\subsubsection{Results}

\paragraph{Improved exploration and smoother training dynamics.}
After replacing the positive-token IS weights with response-level means, all training curves become substantially smoother. The entropy decline slows down (Figure~\ref{fig:main_results}), preventing premature convergence to local optima, and encouraging more diverse and exploratory behavior of the policy. Additionally, the increases in repetition rate, positive clipping ratio, and KL loss are noticeably moderated without the accelerating trends observed in GRPO.
These results provide strong empirical evidence that the original IS ratio design indeed leads to unstable optimization through excessive weighting of high-probability tokens.

\paragraph{Better performance with stable training.}
As shown in Figure~\ref{fig:main_results}, the modified method achieves performance improvements compared with GRPO, demonstrating that stable training does not compromise final performance.

\begin{mybox}{Takeaways for IS Weights of Positive Tokens}
\begin{itemize}[leftmargin=6pt]
    \item The standard PPO-Clip design introduces a \textbf{token-weight mismatch for positive-advantage samples}, where high-probability tokens receive disproportionately large update weights.
    \item This imbalance leads to entropy collapse, and increased repetition, ultimately pushing the policy toward a local optimum and limiting its capacity for continual improvement.
    \item Reweighting positive samples with response-level IS ratios effectively mitigates these issues, confirming the validity of our preliminary analysis and motivating the design of our proposed method in the next section.
\end{itemize}
\end{mybox}

\section{Asymmetric Importance Sampling Policy Optimization}
\label{sec:method}

Based on the analysis in Section~\ref{sec:is_not_important} and~\ref{sec:is_mismatch}, 
we show that token-level IS ratios in GRPO-style OSRL should be understood primarily as token-wise update weights under shared response-level advantages. In addition, the response-level mean-ratio diagnostic in Section~5 shows that removing token-level ratio variation from positive-advantage responses substantially stabilizes training. 
This perspective reveals an asymmetric issue: the original PPO-style ratio weighting is reasonably aligned with negative-advantage responses, but mismatched for positive-advantage responses. 
For $\hat A>0$, tokens with $\pi_\theta \ll \pi_{\mathrm{old}}$ are lagging behind the desired update direction but receive small weights, while tokens with $\pi_\theta \gg \pi_{\mathrm{old}}$ have already been reinforced but receive large weights.

Motivated by this, we propose \textbf{A}symmetric Importance \textbf{S}ampling \textbf{P}olicy \textbf{O}ptimization~(\ourmethod), a simple yet effective approach for clipping and IS ratio computation, following the mismatch analysis in Section~\ref{sec:is_mismatch}. \ourmethod inverts the IS weights of positive samples, aligning their update behavior with that of negative samples. In other words, tokens whose current policy probabilities are lower than the old policy should be assigned higher learning weights, while those with higher probabilities should receive lower weights.

Specifically, the implementation can be divided into three steps:
\paragraph{Step 1: Token Masking.}
We retain the original clipping mechanism in GRPO. The gradient of tokens that satisfy: (1) $r_t^i(\theta) < 1 - \varepsilon_{\text{low}}$ ($\hat{A}_t^i < 0$) or (2) $r_t^i(\theta) > 1 + \varepsilon_{\text{high}}$ ($\hat{A}_t^i > 0$) will be masked in a hard clipping manner.

\paragraph{Step 2: Weight Flipping.}
For tokens with negative advantage values, the \ourmethod ratio is the same as that of GRPO, i.e., $\hat{r}_t^i = r_t^i$.
For tokens with $\hat{A}_t^i > 0$, we use the reciprocal of their IS weights and the \ourmethod ratio is computed as:
\begin{equation}
\hat{r}_t^i = \frac{\textcolor{red}{\pi_{\theta_{\text{old}}}(o_t^i \mid q, o_{<t}^i)} \pi_{\theta}(o_t^i \mid q, o_{<t}^i)}{\textcolor{red}{\operatorname{sg}(\pi_{\theta}^2(o_t^i \mid q, o_{<t}^i))}},
\end{equation}
where $\operatorname{sg}(\cdot)$ denotes stop gradient operation.
The gradient analysis in Appendix~\ref{sec:gradient_analysis} shows that the gradient of positive tokens in \ourmethod is positively correlated with $\frac{1}{\pi_{\theta}}$, indicating that the gradient becomes larger when the probability is lower.

\paragraph{Step 3: Dual Clipping.}
PPO-Clip usually uses a dual-clip mechanism~\citep{Dual-clip-PPO} to handle cases where, for $\hat{A} < 0$, extremely small or large ratios could lead to weight explosion, destabilizing training. Originally, for the $\hat{A} > 0$ region, this problem was naturally avoided by the hard clipping mechanism. However, since we now invert the weights for positive samples, extreme cases shift to the right-hand side of the $\hat{A} > 0$ region (region B in Figure~\ref{fig:is_visualization}(b)). Therefore, when using \ourmethod, positive-sample tokens also require dual-clip. Specifically, this dual-clip is implemented using a \textbf{soft clipping} manner, which only clips the values but retains the gradient.

It is important to note that tokens clipped by dual-clip are fundamentally different from tokens masked in the first step. The latter are blocked because they already have sufficient advantage in the update direction, whereas the former are tokens that lag significantly behind the old policy but need their weight magnitude constrained due to abnormal computation. We still want these tokens to participate in training, so we use the soft clipping in CISPO~\citep{CISPO} for these dual-clipped tokens.

\section{Experiments}
\label{sec:experiments}

\subsection{Setup}
\label{subsec:exp_setup}

\paragraph{Models and Baselines.}

We conduct experiments using \DSOneB~\citep{DeepSeek-R1}, \QFourB, \QEightB, and \QThirtyB\footnote{We use non-thinking mode for Qwen3 models.}~\citep{Qwen3} as the base model.
We compare \ourmethod against several representative baselines: (1) Base Model, (2) GRPO~\citep{GRPO}, (3) DeepScaleR-1.5B~\citep{DeepScaleR}, (4) DeepCoder-1.5B~\citep{DeepCoder}, and (5) Nemotron-1.5B~\citep{ProRL}.

{
\begin{table*}[t]
\centering
\caption{Evaluation results on mathematical benchmarks. The results of \ourmethod are \colorbox{\mycolor}{shaded} and the highest values are \textbf{bolded}.}
\resizebox{1.0\textwidth}{!}{
\begin{tabular}{
L{2.99cm}   
C{0.65cm} C{0.75cm} @{\hspace{0.40cm}}  
C{0.65cm} C{0.75cm} @{\hspace{0.40cm}}  
C{0.65cm} C{0.75cm} @{\hspace{0.40cm}}  
C{0.65cm} C{0.75cm} @{\hspace{0.35cm}}  
C{0.65cm} C{0.75cm} @{\hspace{0.35cm}}  
C{0.65cm} C{0.75cm} @{\hspace{0.35cm}}  
C{0.7cm}   
}
\toprule
\multirow{2}{*}{\textbf{Method}} & \multicolumn{2}{c}{\textbf{AIME24}} & \multicolumn{2}{c}{\textbf{AIME25}} & \multicolumn{2}{c}{\textbf{AMC23}} & \multicolumn{2}{c}{\textbf{MATH-500}} & \multicolumn{2}{c}{\textbf{Minerva}} & \multicolumn{2}{c}{\textbf{Olympiad}} & \multirow{2}{*}{\textbf{Avg.}} \\
\cmidrule(lr){2-3} \cmidrule(lr){4-5} \cmidrule(lr){6-7} \cmidrule(lr){8-9} \cmidrule(lr){10-11} \cmidrule(lr){12-13}
\noalign{\vskip -0.85em}
& {\footnotesize\color{gray}{avg@64}} & {\footnotesize\color{gray}{pass@64}}
& {\footnotesize\color{gray}{avg@64}} & {\footnotesize\color{gray}{pass@64}}
& {\footnotesize\color{gray}{avg@64}} & {\footnotesize\color{gray}{pass@64}}
& {\footnotesize\color{gray}{avg@4}} & {\footnotesize\color{gray}{pass@4}}
& {\footnotesize\color{gray}{avg@8}} & {\footnotesize\color{gray}{pass@8}}
& {\footnotesize\color{gray}{avg@4}} & {\footnotesize\color{gray}{pass@4}}
& \\
\midrule
\textbf{DeepSeek-R1-1.5B} & 30.6 & 80.0 & 23.5 & 63.3 & 70.7 & \textbf{100.0} & 83.6 & 92.4 & 27.6 & 48.2 & 44.6 & 59.4 & 46.8 \\
$\drsh$ GRPO  & 42.1 & 80.0 & 28.6 & 56.7 & 80.3 & 97.5 & 87.6 & \textbf{94.6} & 29.2 & 46.3 & 53.2 & 65.8 & 53.5 \\
$\drsh$ DeepScaleR-1.5B   & 42.0 & \textbf{83.3} & 29.0 & 63.3 & 81.3 & \textbf{100.0} & 87.7 & 93.6 & 30.3 & \textbf{51.1} & 50.7 & 61.0 & 53.5 \\
$\drsh$ Nemotron-1.5B     & 48.0 & 76.7 & 33.1 & 60.0 & 86.1 & 97.5 & \textbf{90.6} & 93.6 & \textbf{35.3} & 47.8 & \textbf{59.2} & 66.8 & 58.7 \\
\rowcolor{cyan!10} $\drsh$ \ourmethod-Math-1.5B & \textbf{49.0} & 80.0 & \textbf{35.1} & \textbf{70.0} & \textbf{87.2} & 95.0 & 90.5 & 94.4 & 35.1 & 50.4 & 58.8 & \textbf{66.9} & \textbf{59.3} \\
\midrule
\textbf{\QFourB} & 23.6 & 56.7 & 18.3 & 63.3 & 67.7 & 95.0 & 84.5 & 92.4 & 41.5 & 56.3 & 54.1 & 66.6 & 48.3 \\ 
$\drsh$ GRPO     & 43.4 & \textbf{83.3} & 35.5 & \textbf{70.0} & 84.3 & \textbf{97.5} & 91.7 & 95.8 & 47.2 & 58.5 & 67.4 & 75.8 & 61.6 \\ 
\rowcolor{cyan!10} $\drsh$ \ourmethod-Math-4B & \textbf{50.5} & \textbf{83.3} & \textbf{40.9} & \textbf{70.0} & \textbf{87.4} & \textbf{97.5} & \textbf{93.4} & \textbf{97.0} & \textbf{49.0} & \textbf{60.7} & \textbf{68.8} & \textbf{78.5} & \textbf{65.0} \\ 
\midrule
\textbf{\QEightB} & 27.0 & 63.3 & 19.1 & 56.7 & 68.9 & 97.5 & 83.6 & 92.4 & 43.9 & 58.1 & 55.7 & 69.6 & 49.7 \\ 
$\drsh$ GRPO      & 50.3 & \textbf{83.3} & 34.1 & 66.7 & 84.1 & 95.0 & 92.7 & 96.0 & 50.2 & \textbf{61.4} & 68.2 & 76.1 & 63.3 \\ 
\rowcolor{cyan!10} $\drsh$ \ourmethod-Math-8B & \textbf{52.4} & \textbf{83.3} & \textbf{38.9} & \textbf{80.0} & \textbf{89.1} & \textbf{97.5} & \textbf{93.6} & \textbf{96.4} & \textbf{50.6} & 60.7 & \textbf{69.8} & \textbf{77.9} & \textbf{65.7} \\
\midrule
\textbf{\QThirtyB} & 30.1 & 70.0 & 19.9 & 56.7 & 74.3 & \textbf{100.0} & 88.4 & 96.0 & 47.7 & 59.6 & 59.6 & 71.2 & 53.3 \\ 
$\drsh$ GRPO       & 59.5 & 80.0 & 43.1 & 73.3 & 91.9 & 97.5 & 95.3 & 97.8 & 51.9 & 61.8 & 71.0 & 79.2 & 68.8 \\ 
\rowcolor{cyan!10} $\drsh$ \ourmethod-Math-30B & \textbf{61.7} & \textbf{90.0} & \textbf{50.3} & \textbf{80.0} & \textbf{94.9} & 97.5 & \textbf{95.7} & \textbf{98.4} & \textbf{54.3} & \textbf{64.3} & \textbf{74.3} & \textbf{81.5} & \textbf{71.8} \\ 
\bottomrule
\end{tabular}%
}
\label{tab:main_math}%
\end{table*}%
}

\begin{table*}[t]
\centering
\caption{Evaluation results on code benchmarks. The results of \ourmethod are \colorbox{\mycolor}{shaded} and the highest values are \textbf{bolded}.}
\resizebox{0.8\textwidth}{!}{
\begin{tabular}{lccccc}
\toprule
\multirow{2}[2]{*}{\textbf{Method}} & \multicolumn{2}{c}{\textbf{LCB v5 (2024.08.01-2025.02.01)}} & \multicolumn{2}{c}{\textbf{LCB v6 (2025.02.01-2025.05.01)}} & \multirow{2}[2]{*}{\textbf{Avg.}} \\
\cmidrule(lr){2-3} \cmidrule(lr){4-5}
  & \multicolumn{1}{c}{\color{gray}{avg@8}}
  & \multicolumn{1}{c}{\color{gray}{pass@8}}
  & \multicolumn{1}{c}{\color{gray}{avg@16}}
  & \multicolumn{1}{c}{\color{gray}{pass@16}}
  & \\
\midrule
\textbf{DeepSeek-R1-1.5B} & 16.7 & 29.0 & 17.2 & 34.4 & 17.0 \\
$\drsh$ GRPO  & 26.0 & 40.5 & 27.6 & 43.5 & 26.8 \\
$\drsh$ DeepCoder-1.5B  & 23.3 & 39.1 & 22.6 & 42.0 & 23.0 \\
$\drsh$ Nemotron-1.5B   & 26.1 & 35.5 & 29.5 & 42.8 & 27.8 \\
\rowcolor{cyan!10} $\drsh$ \ourmethod-Code-1.5B & \textbf{31.5} & \textbf{47.0} & \textbf{30.5} & \textbf{46.0} & \textbf{31.0} \\
\midrule
\textbf{\QFourB} & 23.8 & 35.8 & 24.0 & 35.1 & 23.9 \\
$\drsh$ GRPO & 40.8 & 55.2 & 36.6 & 45.0 & 38.7 \\
\rowcolor{cyan!10} $\drsh$ \ourmethod-Code-4B & \textbf{44.8} & \textbf{58.1} & \textbf{38.3} & \textbf{47.3} & \textbf{41.6} \\
\bottomrule
\end{tabular}%
}
\label{tab:main_code}%
\end{table*}%

\paragraph{Evaluation.}
We evaluate models on both mathematical and coding domains. For math, we use six challenging datasets: AIME24~\citep{AIME24}, AIME25~\citep{AIME25}, AMC23~\citep{AMC23}, MATH-500~\citep{PRM800K}, Minerva Math~\citep{Minerva-Math}, and OlympiadBench~\citep{OlympiadBench}.  
For coding, we adopt LiveCodeBench v5 (2024.08.01–2025.02.01) and v6 (2025.02.01–2025.05.01)~\citep{LiveCodeBench}.  
vLLM~\citep{vLLM} is used for inference with a maximum output length of 32,768 tokens and a temperature of 0.8. We report both avg@K and pass@K for each benchmark.

\paragraph{Implementation Details.}
We implement \ourmethod based on GRPO~\citep{GRPO} with verl~\citep{verl}. The training batch size is set to 64, with a mini-batch size of 16. The learning rate is $1.0 \times 10^{-6}$.
Additional implementation details are provided in Appendix~\ref{app:experimental_details}.

\subsection{Main Results}
\label{subsec:exp_main}

The results in Table~\ref{tab:main_math} and Table~\ref{tab:main_code} show that, when using \DSOneB as the base model, \ourmethod improves upon the base model by 12.5\% and 14.0\% on math and coding tasks, respectively. Moreover, \ourmethod consistently outperforms GRPO and several strong OSRL methods averaged across all evaluated benchmarks, demonstrating the effectiveness of \ourmethod.
On larger dense and MoE models, \ourmethod still exceeds the baselines by a large margin, demonstrating the effectiveness across different model sizes.
As shown in Table~\ref{tab:main_ood}, \ourmethod still outperforms the baseline method on OOD tasks.

\subsection{Analysis}
\label{subsec:analysis}

As shown in Figure~\ref{fig:main_results} and Table~\ref{tab:main_math}, compared with GRPO, \ourmethod yields trends that align with our earlier findings, but with even stronger improvements. Specifically, entropy decreases more gradually, the repetition rate and positive token clipping ratio grow more slowly, and all metrics eventually stabilize, showing characteristics of healthy convergence as discussed in Appendix~\ref{app:distinguish_convergence_or_local_optima}. More training dynamics results are shown in Appendix~\ref{app:exp_results}.

More importantly, since entropy declines more smoothly and remains at a higher level, the model avoids premature collapse and continues learning effectively. As training progresses, performance steadily improves, significantly surpassing the best results achieved by GRPO-based training.

It is also noteworthy that in the early training stages, models trained with \ourmethod exhibit slightly slower reward improvement (Figure~\ref{fig:training_dynamics_qwen3_4b}(a) and~\ref{fig:training_dynamics_qwen3_8b}(a)) compared to other variants. This occurs because inverting positive-sample weights reduces the overall average IS weight, leading to slower initial fitting of positive samples. However, as training continues, the performance not only catches up but ultimately surpasses the other approaches.

\section{Conclusion}
\label{sec:conclusion}

In this paper, we identify a fundamental role shift of IS in GRPO-style OSRL: under shared response-level advantages, token-level IS ratios primarily act as token-wise update weights rather than distribution-correction terms. We further reveal a weight mismatch for positive-advantage tokens, where already-confident tokens are over-amplified while lagging tokens are under-updated, leading to entropy collapse and unstable training. To address this, we propose \ourmethod, which flips the IS ratios of positive tokens and incorporates a soft dual-clipping mechanism to stabilize OSRL for LLMs. Experimental results across mathematical and coding domains demonstrate that our method effectively alleviates token-level weight mismatch, mitigates entropy collapse, and improves training stability, leading to superior model performance than the baselines.

\paragraph{Limitations and Discussion.}
Our investigation primarily targets the importance sampling mechanism within the widely-adopted GRPO-based methods, which have recently emerged as a popular paradigm for efficient LLM post-training.
It is a promising direction to extend our investigation to other RL paradigms, e.g., PPO with token-level IS ratios and advantages. We hope our work serves as a foundation for these broader explorations.

\bibliographystyle{plain}
\bibliography{ref}


\newpage
\appendix

\section{Theoretical Analysis}

\subsection{Proof of Proposition 4.1} \label{sec:proof}
\textbf{Proposition~\ref{prop}} (IS ratio as token-wise gradient weighting in GRPO)\textbf{.}\emph{
Consider the GRPO objective in Eq.~(\ref{eq:grpo_loss}). For each token $o_t^i$, define the active clipping mask $m_t^i=\begin{cases}1, & A_t^i \ge 0 {\rm{\ and\ }} r_t^i \le 1+\epsilon, \\
1, & A_t^i < 0 {\rm{\ and\ }} r_t^i \ge 1-\epsilon, \\
0, & {\rm{otherwise}}
\end{cases}$
and the detached token weight $w_t^i={\rm{sg}}(m_t^i r_t^i)$, where $\rm{sg}(\cdot)$ denotes the stop-gradient operator. Consider the weighted log-likelihood objective in Eq.~(\ref{eq:wll_loss_app}), then we have that $\nabla_\theta \mathcal J_{\mathrm{GRPO}}(\theta)=\nabla_\theta\mathcal J_{\mathrm{WLL}}(\theta)$.
\begin{equation}
\mathcal{J}_{\text{WLL}}(\theta) = \mathbb{E}_{q \sim \mathcal{D}, \{o^i\}_{i=1}^G \sim \pi_{\theta_{\text{old}}}(\cdot \mid q)} 
\left[ \frac{1}{G} \sum_{i=1}^G \frac{1}{|o^i|} \sum_{t=1}^{|o^i|} \bigg( w_t^i A_t^i \log \pi_{\theta}(o_t^i | q,o_{<t}^i) - \beta \mathbb{D}_{\text{KL}}(\pi_\theta \| \pi_{\text{ref}}) \bigg) \right].
\label{eq:wll_loss_app}
\end{equation}
}
\begin{proof}
We prove in three cases:

Case 1: $A_t^i \ge 0 {\rm{\ and\ }} r_t^i > 1+\epsilon$. We have \begin{equation}
\mathcal{J}_{\text{WLL}}(\theta) = \mathbb{E}_{q \sim \mathcal{D}, \{o^i\}_{i=1}^G \sim \pi_{\theta_{\text{old}}}(\cdot \mid q)} 
\left[ \frac{1}{G} \sum_{i=1}^G \frac{1}{|o^i|} \sum_{t=1}^{|o^i|} \bigg( - \beta \mathbb{D}_{\text{KL}}(\pi_\theta \| \pi_{\text{ref}}) \bigg) \right].
\end{equation}
Also, we have

\begin{equation}
\begin{aligned}
\mathcal{J}_{\text{GRPO}}(\theta) = \ \mathbb{E}_{q \sim \mathcal{D}, \{o^i\}_{i=1}^G \sim \pi_{\theta_{\text{old}}}(\cdot \mid q)} 
&\left[ \frac{1}{G} \sum_{i=1}^G \frac{1}{|o^i|} \sum_{t=1}^{|o^i|} \bigg( (1+\varepsilon ) \hat{A}_t^i- \beta \mathbb{D}_{\text{KL}}(\pi_\theta \| \pi_{\text{ref}}) \bigg) \right].
\end{aligned}
\end{equation}

Thus, we have $\nabla_\theta \mathcal J_{\mathrm{GRPO}}(\theta)=\nabla_\theta\mathcal J_{\mathrm{WLL}}(\theta)$.

Case 2: $A_t^i < 0 {\rm{\ and\ }} r_t^i < 1-\epsilon$. We have \begin{equation}
\mathcal{J}_{\text{WLL}}(\theta) = \mathbb{E}_{q \sim \mathcal{D}, \{o^i\}_{i=1}^G \sim \pi_{\theta_{\text{old}}}(\cdot \mid q)} 
\left[ \frac{1}{G} \sum_{i=1}^G \frac{1}{|o^i|} \sum_{t=1}^{|o^i|} \bigg( - \beta \mathbb{D}_{\text{KL}}(\pi_\theta \| \pi_{\text{ref}}) \bigg) \right].
\end{equation}
Also, we have 

\begin{equation}
\begin{aligned}
\mathcal{J}_{\text{GRPO}}(\theta) = \ \mathbb{E}_{q \sim \mathcal{D}, \{o^i\}_{i=1}^G \sim \pi_{\theta_{\text{old}}}(\cdot \mid q)} 
&\left[ \frac{1}{G} \sum_{i=1}^G \frac{1}{|o^i|} \sum_{t=1}^{|o^i|} \bigg( (1-\varepsilon ) \hat{A}_t^i- \beta \mathbb{D}_{\text{KL}}(\pi_\theta \| \pi_{\text{ref}}) \bigg) \right].
\end{aligned}
\end{equation}

Thus, we have $\nabla_\theta \mathcal J_{\mathrm{GRPO}}(\theta)=\nabla_\theta\mathcal J_{\mathrm{WLL}}(\theta)$.

Case 3: Otherwise, we have
\begin{equation}
\begin{aligned}
\mathcal{J}_{\text{WLL}}(\theta)
= {} & \mathbb{E}_{q \sim \mathcal{D}, \{o^i\}_{i=1}^G \sim \pi_{\theta_{\text{old}}}(\cdot \mid q)}
\Bigg[
\frac{1}{G} \sum_{i=1}^G \frac{1}{|o^i|} \sum_{t=1}^{|o^i|}
\bigg(
sg\!\left(
\frac{\pi_{\theta}(o_t^i | q,o_{<t}^i)}
{\pi_{old}(o_t^i | q,o_{<t}^i)}
\right) A_t^i \log \pi_{\theta}(o_t^i | q,o_{<t}^i)
\\
& \qquad\qquad\qquad\qquad
- \beta \mathbb{D}_{\text{KL}}(\pi_\theta \| \pi_{\text{ref}})
\bigg)
\Bigg].
\end{aligned}
\end{equation}

\begin{equation}
\begin{aligned}
\mathcal{J}_{\text{GRPO}}(\theta) = \ \mathbb{E}_{q \sim \mathcal{D}, \{o^i\}_{i=1}^G \sim \pi_{\theta_{\text{old}}}(\cdot \mid q)} 
 \left[ \frac{1}{G} \sum_{i=1}^G \frac{1}{|o^i|} \sum_{t=1}^{|o^i|} \bigg(  \frac{\pi_{\theta}(o_t^i | q,o_{<t}^i)}{\pi_{old}(o_t^i | q,o_{<t}^i)} \hat{A}_t^i - \beta \mathbb{D}_{\text{KL}}(\pi_\theta \| \pi_{\text{ref}}) \bigg) \right].
\end{aligned}
\end{equation}

Thus, we have $\nabla_\theta \mathcal J_{\mathrm{GRPO}}(\theta)=\nabla_\theta\mathcal J_{\mathrm{WLL}}(\theta)$.

\end{proof}

\subsection{Gradient Analysis}
\label{sec:gradient_analysis}

The gradient of original GRPO is as follows:
\begin{equation}
\begin{aligned}
\nabla_\theta \mathcal{J}(\theta) &= \nabla_\theta \mathbb{E}_{q \sim \mathcal{D}, \{o^i\}_{i=1}^G \sim \pi_{\theta_{\text{old}}}(\cdot \mid q)} \left[ \frac{1}{G} \sum_{i=1}^G r_t^i(\theta) \hat{A}_t^i \right] \\
&= \mathbb{E}_{q \sim \mathcal{D}, \{o^i\}_{i=1}^G \sim \pi_{\theta_{\text{old}}}(\cdot \mid q)} \left[ \frac{1}{G} \sum_{i=1}^G \frac{\nabla_\theta \pi_{\theta}(o_t^i \mid q, o_{<t}^i)}{\pi_{\theta_{\text{old}}}(o_t^i \mid q, o_{<t}^i)} \hat{A}_t^i \right] \\
&= \mathbb{E}_{q \sim \mathcal{D}, \{o^i\}_{i=1}^G \sim \pi_{\theta_{\text{old}}}(\cdot \mid q)} \left[ \frac{1}{G} \sum_{i=1}^G \frac{\pi_{\theta}(o_t^i \mid q, o_{<t}^i)}{\pi_{\theta_{\text{old}}}(o_t^i \mid q, o_{<t}^i)} \nabla_\theta \log\pi_{\theta}(o_t^i \mid q, o_{<t}^i) \hat{A}_t^i \right] \\
\end{aligned}
\label{eq:grpo_gradient}
\end{equation}
where $\pi_{\theta}$ denotes $\pi_{\theta}(o_t^i \mid q, o_{<t}^i)$ and $\pi_{\theta_{\text{old}}}$ denotes $\pi_{\theta_{\text{old}}}(o_t^i \mid q, o_{<t}^i)$.
Then, we derive the gradient of positive tokens in \ourmethod as follows:
\begin{equation}
\begin{aligned}
\nabla_\theta \mathcal{J}(\theta) &= \nabla_\theta \mathbb{E}_{q \sim \mathcal{D}, \{o^i\}_{i=1}^G \sim \pi_{\theta_{\text{old}}}(\cdot \mid q)} \left[ \frac{1}{G} \sum_{i=1}^G \hat{r}_t^i(\theta) \hat{A}_t^i \right] \\
&= \mathbb{E}_{q \sim \mathcal{D}, \{o^i\}_{i=1}^G \sim \pi_{\theta_{\text{old}}}(\cdot \mid q)} \left[ \frac{1}{G} \sum_{i=1}^G \frac{\pi_{\theta_{\text{old}}}(o_t^i \mid q, o_{<t}^i) \nabla_\theta \pi_{\theta}(o_t^i \mid q, o_{<t}^i)}{\operatorname{sg}(\pi_{\theta}^2(o_t^i \mid q, o_{<t}^i))} \hat{A}_t^i \right] \\
&= \mathbb{E}_{q \sim \mathcal{D}, \{o^i\}_{i=1}^G \sim \pi_{\theta_{\text{old}}}(\cdot \mid q)} \left[ \frac{1}{G} \sum_{i=1}^G \frac{\pi_{\theta_{\text{old}}}(o_t^i \mid q, o_{<t}^i) \pi_{\theta}(o_t^i \mid q, o_{<t}^i)}{\pi_{\theta}^2(o_t^i \mid q, o_{<t}^i)} \nabla_\theta \log\pi_{\theta}(o_t^i \mid q, o_{<t}^i) \hat{A}_t^i \right] \\
&= \mathbb{E}_{q \sim \mathcal{D}, \{o^i\}_{i=1}^G \sim \pi_{\theta_{\text{old}}}(\cdot \mid q)} \left[ \frac{1}{G} \sum_{i=1}^G \textcolor{red}{\frac{\pi_{\theta_{\text{old}}}(o_t^i \mid q, o_{<t}^i)}{\pi_{\theta}(o_t^i \mid q, o_{<t}^i)}} \nabla_\theta \log\pi_{\theta}(o_t^i \mid q, o_{<t}^i) \hat{A}_t^i \right] \\
\end{aligned}
\label{eq:aspo_gradient}
\end{equation}
It is worth noting that the gradient of positive tokens in~\eqref{eq:aspo_gradient} differs from the original gradient of GRPO in~\eqref{eq:grpo_gradient} at the highlighted red term.
From the above derivation, we can observe that the gradient of \ourmethod is positively correlated with $\frac{1}{\pi_{\theta}}$, indicating that the gradient becomes larger when the probability of a token is lower.

\section{Experimental Details}
\label{app:experimental_details}

\paragraph{Baselines.}

We conduct experiments using \DSOneB~\citep{DeepSeek-R1}, \QFourB, \QEightB, and \QThirtyB~\citep{Qwen3} as the base model and compare \ourmethod with the following baselines:
\begin{itemize}
    \item \textbf{Base Model}: The original model without any RL fine-tuning.
    \item \textbf{GRPO}~\citep{GRPO}: A strong OSRL algorithm with token-level loss.
    \item \textbf{DeepScaleR-1.5B}~\citep{DeepScaleR}: A 1.5B model trained for mathematical reasoning with iterative context-length expansion.
    \item \textbf{DeepCoder-1.5B}~\citep{DeepCoder}: A 1.5B model trained on code datasets using similar context expansion strategies as DeepScaleR.
    \item \textbf{Nemotron-1.5B}~\citep{ProRL}: A strong 1.5B reasoning model with reference policy resetting.
\end{itemize}

\paragraph{Evaluation.}
We follow standard practice~\citep{ProRL, Archer} and report both \textit{avg@K} and \textit{pass@K} metrics.
For benchmarks with fewer samples (AIME24/25 and AMC23), we set $K=64$. For LiveCodeBench v6, we use $K=16$; for LiveCodeBench v5 and Minerva Math, $K=8$; and for MATH-500 and OlympiadBench, $K=4$.
To ensure fair and accurate evaluation, we adopt the official verification functions from both DeepScaleR and Math-Verify\footnote{\url{https://github.com/huggingface/Math-Verify}} for mathematical problems, following the protocol in~\citet{Skywork-OR1}.

\paragraph{Implementation Details.}
For training data, we use a mixture of DeepScaleR-Preview-Dataset~\citep{DeepScaleR}, Skywork-OR1-RL-Data~\citep{Skywork-OR1}, and DAPO-Math-17K~\citep{DAPO} for \DSOneB on mathematical tasks.
For \QFourB, \QEightB, and \QThirtyB, we directly use DAPO-Math-17K for training.
For coding, we employ DeepCoder~\citep{DeepCoder}, CodeContests~\citep{CodeContests}, and CodeForces~\citep{CodeForces} datasets.
All datasets are cleaned and filtered following the preprocessing protocol of~\citet{Archer}.
After filtering, the mathematical dataset contains 70.8k samples, while the coding dataset contains 8.9k samples.
The clipping ranges of GRPO and \ourmethod are set to $\varepsilon = 0.2$. The KL divergence is incorporated as an explicit loss term (k3 estimator) with coefficient $\beta=0.001$, and all baselines use the same KL implementation to ensure that observed differences in entropy and stability are not caused by KL.
For each prompt, 16 responses are sampled with a temperature of 1.0. We use a max response length of 32,768 tokens for \DSOneB, 8,192 for \QFourB and \QThirtyB, and 6,144 for \QEightB.
Experiments of 1.5B, 4B, and 8B LLMs are conducted on 8 NVIDIA H800 GPUs and \QThirtyB is trained with 32 NVIDIA H800 GPUs.

\section{Additional Experimental Results}
\label{app:exp_results}

\begin{figure*}[!h]
\centering
\includegraphics[width=1\linewidth]{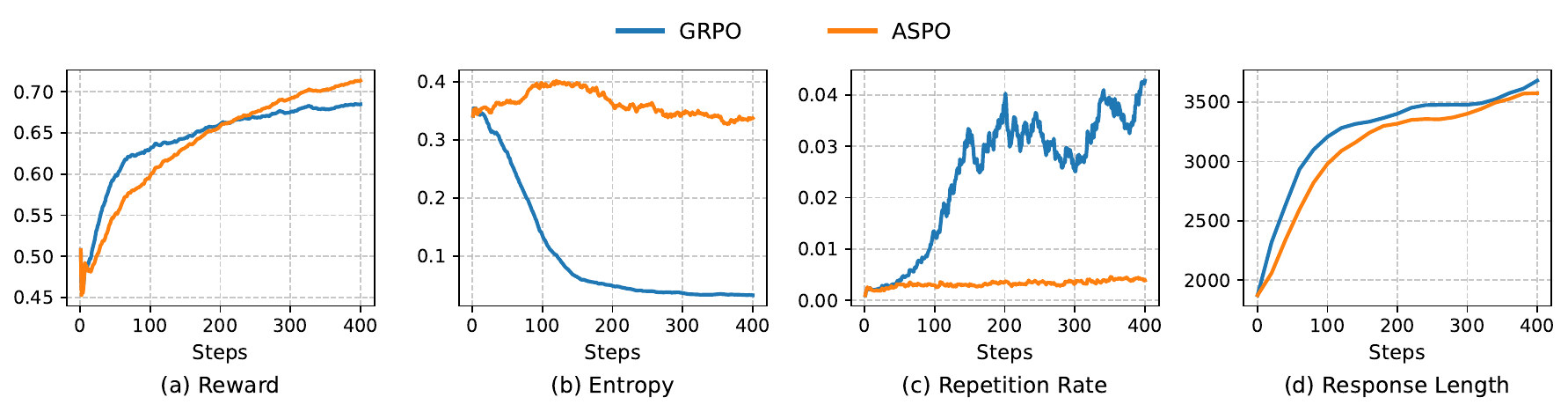}
\caption{The training dynamics curves of all methods on \QFourB, including (a) training reward, (b) model entropy, (c) repetition rate, and (d) response length. The curves are smoothed with EMA for better visualization.}
\label{fig:training_dynamics_qwen3_4b}
\end{figure*}

\begin{figure*}[!h]
\centering
\includegraphics[width=1\linewidth]{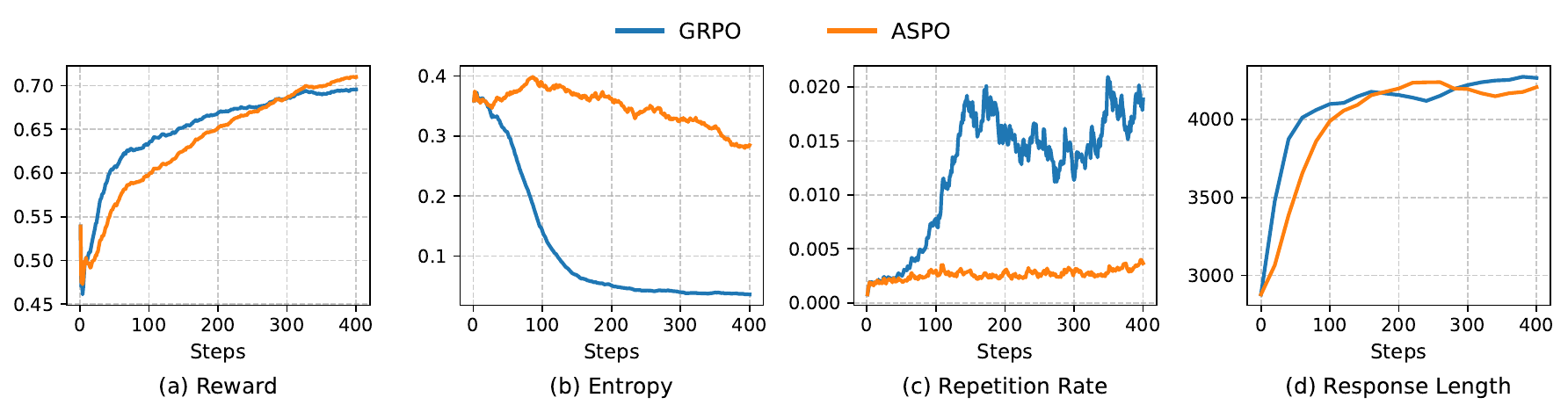}
\caption{The training dynamics curves of all methods on \QEightB, including (a) training reward, (b) model entropy, (c) repetition rate, and (d) response length. The curves are smoothed with EMA for better visualization.}
\label{fig:training_dynamics_qwen3_8b}
\end{figure*}

\begin{figure*}[!h]
\centering
\includegraphics[width=1\linewidth]{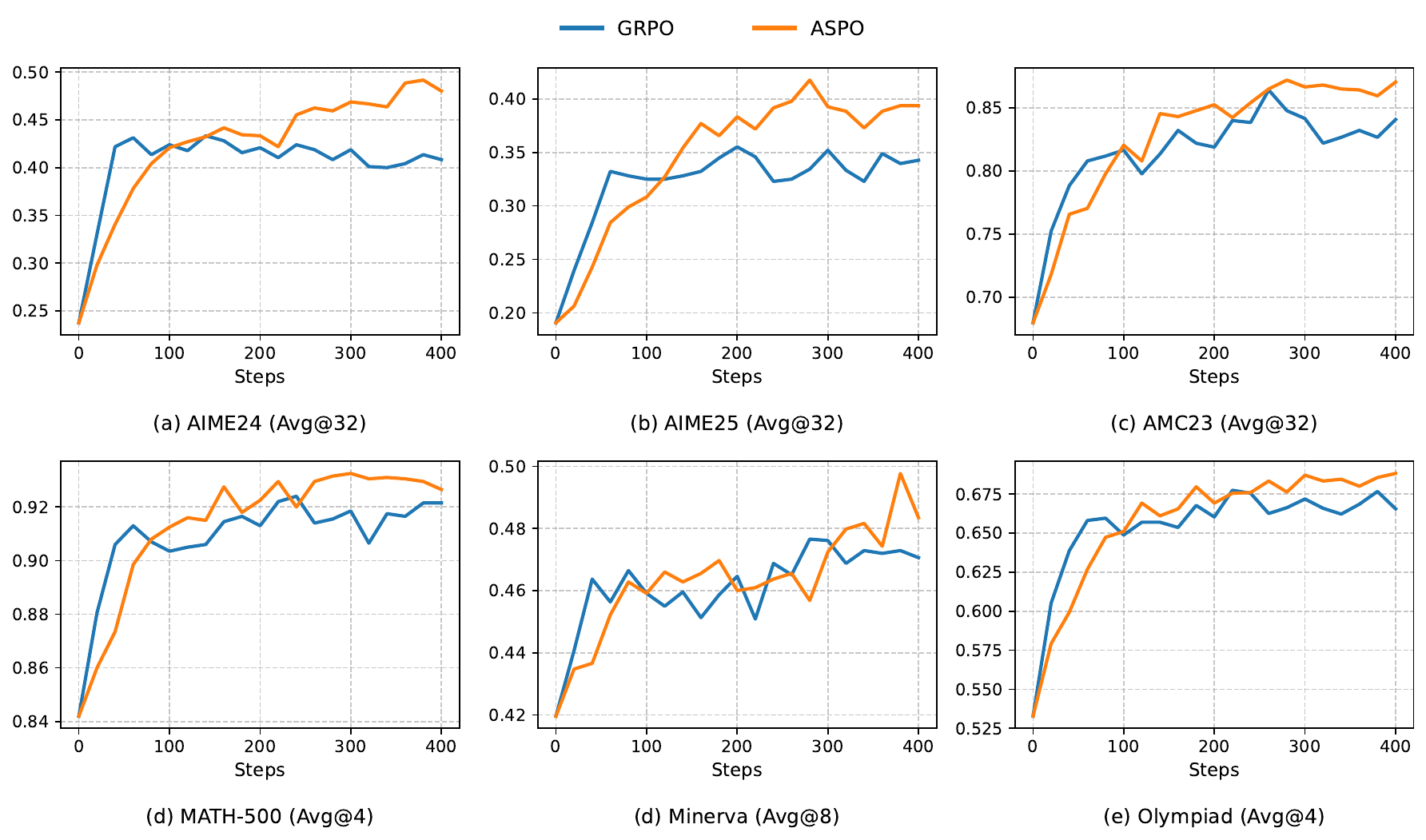}
\caption{The test curves of all methods trained with \QFourB on six mathematical benchmarks.}
\label{fig:all_test_curves_qwen3_4b}
\end{figure*}

We present the training dynamics and test performance curves on mathematical tasks in Figure~\ref{fig:training_dynamics_qwen3_4b},~\ref{fig:training_dynamics_qwen3_8b}, and~\ref{fig:all_test_curves_qwen3_4b}.
As shown in Figures~\ref{fig:training_dynamics_qwen3_4b} and~\ref{fig:training_dynamics_qwen3_8b}, although \ourmethod exhibits slightly slower learning during the initial training phase compared to GRPO, it maintains more stable model entropy and consistently lower repetition rate throughout training. These results demonstrate the superior training stability of \ourmethod.

\begin{figure*}[!h]
\centering
\includegraphics[width=1\linewidth]{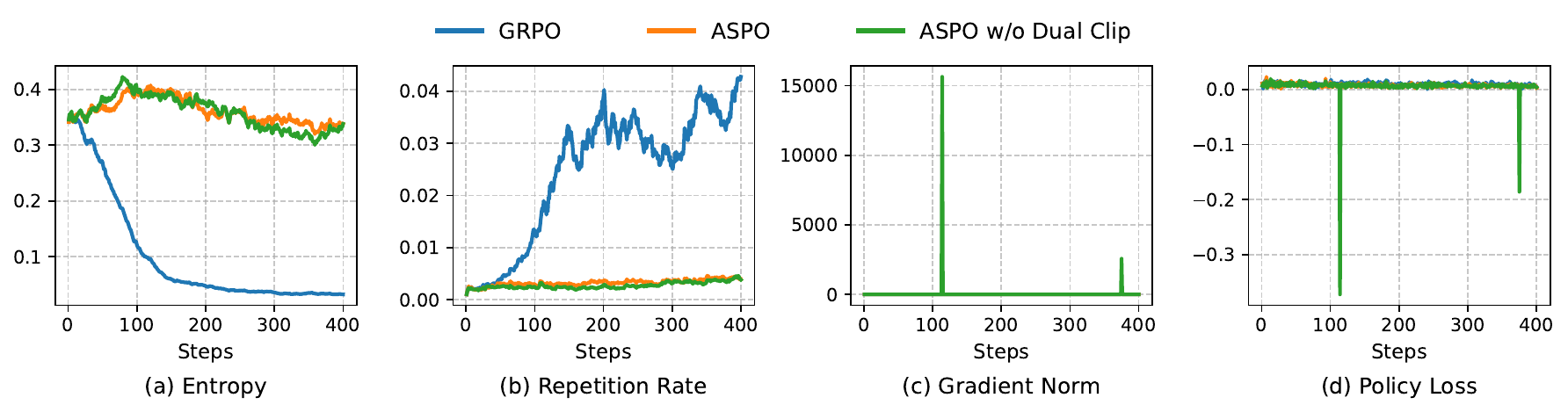}
\caption{Ablation on dual clipping. The training dynamics curves of \ourmethod with and without dual clipping on \QFourB, including (a) entropy, (b) repetition rate, (c) gradient norm, and (d) policy loss. Without dual clipping, the gradient norm and policy loss exhibit explosive spikes during training, demonstrating that dual clipping is essential for stabilizing training in extreme cases. The curves are smoothed with EMA for better visualization.}
\label{fig:ablation_wodualclip}
\end{figure*}

\paragraph{Ablation on Dual Clipping.}
Weight Flipping is the core mechanism of \ourmethod, while Dual Clipping is a stabilizer introduced because, after inversion, extreme positive-token ratios shift to the side where their magnitude also needs to be bounded. As shown in Figure~\ref{fig:ablation_wodualclip}, without Dual Clipping, the gradient norm exhibits explosive spikes during training, while \ourmethod with Dual Clipping maintains stable gradient dynamics. This confirms that Dual Clipping is essential for stabilizing training in extreme cases, complementing the Weight Flipping mechanism.

\begin{table*}[!h]
\centering
\caption{Evaluation results on OOD benchmarks. The results of \ourmethod are \colorbox{\mycolor}{shaded} and the highest values are \textbf{bolded}.}
\resizebox{0.8\textwidth}{!}{
\begin{tabular}{lcccc}
\toprule
\textbf{Method} & \textbf{ARC-Challenge} & \textbf{GPQA-Diamond} & \textbf{MMLU-Pro} & \textbf{Avg.} \\
\midrule
\textbf{\QFourB} & 74.7 & 14.3 & 31.5 & 40.2 \\
$\drsh$ GRPO & \textbf{90.1} & 29.2 & 49.6 & 56.3 \\
\rowcolor{cyan!10} $\drsh$ \ourmethod-Math-4B & 87.6 & \textbf{37.3} & \textbf{53.5} & \textbf{59.5} \\
\midrule
\textbf{\QEightB} & 52.9 & 8.6 & 29.6 & 30.4 \\
$\drsh$ GRPO & 62.4 & 33.0 & 42.9 & 40.6 \\ 
\rowcolor{cyan!10} $\drsh$ \ourmethod-Math-8B & \textbf{88.6} & \textbf{56.3} & \textbf{57.4} & \textbf{59.8} \\ 
\bottomrule
\end{tabular}%
}
\label{tab:main_ood}%
\end{table*}%

\begin{table*}[!h]
\centering
\caption{Evaluation results with \QFourB. The highest values are \textbf{bolded}.}
\resizebox{0.90\textwidth}{!}{
\begin{tabular}{l@{\hspace{4pt}} NNNNNNNNNNNN @{\hspace{5pt}}c}
\toprule
{\textbf{Method}} & \textbf{AIME24} & \textbf{AIME25} & \textbf{AMC23} & \textbf{MATH-500} & \textbf{Minerva} & \textbf{Olympiad} & {\textbf{Avg.}} \\
\midrule
GRPO            & 43.4 & 35.5 & 84.3 & 91.7 & 47.2 & \textbf{67.4} & 61.6 \\ 
GRPO w/o IS & 46.4 & 36.3 & \textbf{88.9} & \textbf{92.9} & \textbf{49.3} & 67.2 & 63.5 \\ 
GRPO w/ Pos. Resp. IS Mean & \textbf{46.5} & \textbf{42.3} & 87.4 & 92.0 & 48.6 & 67.1 & \textbf{64.0} \\ 
\bottomrule
\end{tabular}%
}
\label{tab:abla_is_new}%
\end{table*}%

\section{How to distinguish between ``healthy convergence'' and ``local optima''?}
\label{app:distinguish_convergence_or_local_optima}

To better contextualize our analysis, we introduce an important concept in RL training: how to distinguish between \textbf{healthy convergence} and \textbf{local optima}.

In RL training, healthy convergence typically exhibits the following characteristics:
\begin{itemize}
    \item \textbf{Entropy} decreases gradually from a relatively high initial value and stabilizes at a small but positive level, indicating that the policy becomes more deterministic while retaining moderate exploration.
    \item The \textbf{reward curve} increases steadily and eventually plateaus at a stable high value.
    \item \textbf{Clip ratios} and \textbf{KL divergence loss} remain stable during later training stages, without drastic fluctuations.
\end{itemize}

In contrast, when the training becomes trapped in a local optimum, the model enters a self-reinforcing \textbf{policy-data distribution loop}, characterized by:
\begin{itemize}
    \item \textbf{Entropy} collapses rapidly toward zero.
    \item The \textbf{reward curve} stagnates with no further improvement.
    \item Persistently high \textbf{clip ratios} without meaningful policy updates.
\end{itemize}

When training GRPO-based approaches, we may observe this phenomenon: in late-stage training, entropy collapses, repetition rates spike, clip ratios surge, and the model's test performance begins to degrade, indicating convergence to a local optimum.  

In early training, \textbf{moderate entropy reduction} and a \textbf{gradual increase in clip ratio} are expected. If accompanied by an \textbf{increasing rewards}, it indicates that the policy is learning effectively.
However, when later stages exhibit an abrupt entropy drop, reward stagnation, and persistently high clip ratios without further progress, it clearly signals that the model has fallen into a \textbf{local optimum}.  

The underlying cause lies in the \textbf{token-level weight mismatch for positive samples in PPO-Clip} identified in Section~\ref{sec:is_mismatch}, which drives the model to \textbf{overfit} certain high-probability tokens, eventually leading to entropy collapse and training degradation.

\end{document}